\documentclass[journal]{IEEEtran}

\IEEEoverridecommandlockouts                              

\usepackage[utf8]{inputenc}
\usepackage{xcolor}
\usepackage{soul}
\usepackage{cite}
\usepackage{amsmath,amssymb,amsfonts}
\usepackage{algorithmic}
\usepackage{graphicx}
\usepackage{textcomp}
\usepackage{xcolor}
\usepackage{txfonts}
\usepackage{gensymb}
\usepackage{float}
\usepackage[justification=centering]{caption}
\def\BibTeX{{\rm B\kern-.05em{\sc i\kern-.025em b}\kern-.08em
    T\kern-.1667em\lower.7ex\hbox{E}\kern-.125emX}}
\usepackage{subcaption}
\usepackage[colorlinks]{hyperref}

\usepackage{pdfpages}

\begin{document}

\title{Analog Twin Framework for Human and AI Supervisory Control and Teleoperation of Robots}

\author{Nazish Tahir, \and Ramviyas Parasuraman
\thanks{The authors are with the Heterogeneous Robotics Research Lab (HeRoLab), School of Computing, University of Georgia, Athens, GA 30602, USA. Authors email: {\tt\small \{nazish.tahir,ramviyas\}@uga.edu}}}

\makeatletter
\def\ps@IEEEtitlepagestyle{%
  \def\@oddhead{\mycopyrightnotice}%
  \def\@oddfoot{\hbox{}\@IEEEheaderstyle\leftmark\hfil\thepage}\relax
  \def\@evenhead{\@IEEEheaderstyle\thepage\hfil\leftmark\hbox{}}\relax
  \def\@evenfoot{}%
}
\def\mycopyrightnotice{%
  \begin{minipage}{\textwidth}
  \centering \scriptsize
  This article has been accepted for publication in the IEEE Transactions on Systems, Man, and Cybernetics: Systems with DOI: 10.1109/TSMC.2022.3216206. \\ Copyright~\copyright~20XX IEEE.  Personal use of this material is permitted.  Permission from IEEE must be obtained for all other uses, in any current or future media, including reprinting/republishing this material for advertising or promotional purposes, creating new collective works, for resale or redistribution to servers or lists, or reuse of any copyrighted component of this work in other works.
  \end{minipage}
}
\makeatother

\maketitle

\begin{abstract}
Resource-constrained mobile robots that lack the capability to be completely autonomous can rely on a human or AI supervisor acting at a remote site (e.g., control station or cloud) for their control. Such a supervised autonomy or cloud-based control of a robot poses high networking and computing capabilities requirements at both sites, which are not easy to achieve. This paper introduces and analyzes a new analog twin framework by synchronizing mobility between two mobile robots, where one robot acts as an analog twin to the other robot. 
We devise a novel priority-based supervised bilateral teleoperation strategy for goal navigation tasks to validate the proposed framework. 
The practical implementation of a supervised control strategy on this framework entails a mobile robot system divided into a Master-Client scheme over a communication channel where the Client robot resides on the site of operation guided by the Master robot through an agent (human or AI) from a remote location.
The Master robot controls the Client robot with its autonomous navigation algorithm, which reacts to the predictive force received from the Client robot. 
We analyze the proposed strategy in terms of network performance (throughput and delay), task performance (tracking error and goal reach accuracy), and computing efficiency (memory and CPU utilization). Extensive simulations and real-world experiments demonstrate the method's novelty, flexibility, and versatility in realizing reactive planning applications with remote computational offloading capabilities compared to conventional offloading schemes.

\end{abstract}


\IEEEpeerreviewmaketitle

\section{Introduction}

\IEEEPARstart {C}{yber-Physical} Systems (CPS) encompass robots, autonomous vehicles, smart grids, and other such complex systems. Recent developments in Industry 4.0 Technology and the future Industry 5.0 innovations are driven through the advancements in CPS and robotic systems, with the concept of Digital Twin (DT) playing a pivotal role in it \cite{kousi2019digital}. 
Digital twin includes two components; a physical device and a digital replica. The digital module is the closest resemblance to the physical entity in a software/simulator environment.
Here, both modules exchange information collected via network links, potentially including a cloud computing module. Besides being used as a prototype in digital space representing a physical entity in the real world, DT also allows for commanding control and remote supervision of tasks. 

In robotics research and development, DT provides the flexibility of many teams working independently and simultaneously without access to the physical robot, with the added benefit of safety for experimenting in a digital environment. Although DT demands high fidelity (increasing the requirement for data storage and transmission), it creates an opportunity to visualize the active state of the robot and test the process or models in dedicated offline training phases before deploying the models on real robots. However, DT is currently only used for design, simulations, modeling, or verification before deploying the models onto real robots. Currently, DT is limited in evaluating algorithms and models in synchronization with the actual robot operations \cite{stkaczek2021digital}.

\begin{figure}[t]
    \centering
    \begin{center}
    \includegraphics[width=.48\textwidth]{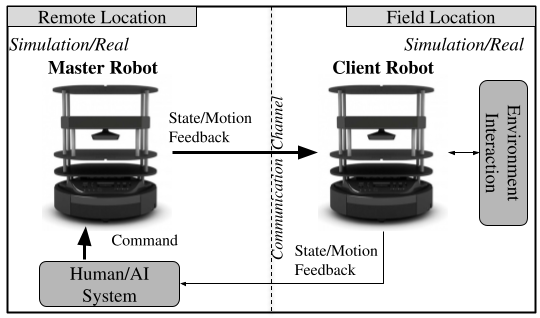}
    \caption{An overview of the Analog Twin framework with a Master-Client architecture for mobile robot teleoperation.}
    \label{fig:overview}
    \vspace{-4mm}
\end{center}
\end{figure} 

We address this gap by proposing an analog twin (AT) framework, focusing on a mobile robot system. Our analog twin framework provides a unique way to verify robot algorithms and create new algorithms that can enable task or computation collaboration between the real robot and its twin robot. In contrast to the DT framework, the robots (whether physical or simulated) are synchronized in their mobility and actions in the analog twin, while digitally twin robots are typically asynchronous. The proposed AT framework can accommodate both simulated and real robots on both sides of the synchronized system. Fig. \ref{fig:overview} depicts a supervised teleoperation application of a mobile robot operating in a real-world environment (field location) in collaboration with a remotely located analog twinned robot using the AT framework.

Teleoperation and navigation are essential control tasks for mobile robots in real-world applications, including urban search and rescue scenarios \cite{kohlbrecher2013hector,Nourbakhsh2005}, cooperative search and transportation \cite{luo2020asymptotic,Sheridan1995}.
In most cases, a simple processor on the robot's most basic setup can only handle the sensor data from the cameras, laser scanners, actuators, etc. On the other hand, real-time applications of robotics, such as artificial vision and object recognition and tracking, are resource-intensive with stringent requirements on the processing, modeling, and training of a constant stream of video data about the robot's surroundings. Their resources may become insufficient to do such resource-intensive operations, resulting in reduced computational performance.
Such applications can be significantly benefited by Cloud robotics \cite{chinchali2021network,kehoe2015survey}, by allowing cooperative computation with robots and sharing resources among themselves or offloading calculations to a resource-rich remote server/cloud. The processed outputs are delivered back to the robots for incorporation into the overall computation since the expensive computations are conducted either by robot-robot or robot-cloud collaborative computing \cite{manzi2016use}.

This paper applies an analog twin framework to a mobile robot system, focusing on a supervised teleoperation and navigation task. Our objective is to enhance robot resource efficiency, connectivity, and usage by offloading computationally heavy operations to remote computers/servers. We aim to improve the following key performance metrics that are relevant to cloud-connected robots \cite{chen2018study}: computing workload, network latency, and data throughput without affecting the control task performance. 

The main contributions of this paper are the following:
\begin{itemize}
    \item We propose a new framework termed Analog Twin, exploring the novel idea of synchronized (supervised) control of mobile robots through a networked system. The new framework enables us to utilize the capabilities of a fully autonomous robot to operate another robot for algorithmic collaboration and verification remotely. 
    \item We implement a new priority-based bilateral teleoperation strategy on the AT framework to share the networking and computing resources between two robots for realizing supervised control and navigation.
    \item We construct a collaborative robotics scenario in which the less computational field robot offloads its intensive computational tasks to a remote (or cloud) server that handles the calculations, thereby reducing the field robot's computational overhead cost. 
    \item We validate the proposed system in various situations with increasing complexities through extensive simulation experiments with multiple obstacle settings and demonstrate the approach with real-world robot (hardware) demonstrations. We look at each instance's navigation, latency, communication, and computation performances.
    \item We analyze the performances compared to conventional onboard computation and remote offloading schemes.
\end{itemize}

The proposed strategy enhances remotely-offloaded robot navigation tasks by achieving improved computing and network performances without losing the bilateral system's tracking (control) performance. 
The proposed contributions are critical to enhancing the CPS and Robotics research in the field and industrial applications by creating a new class of Analog Twin-based telerobotics and cyber-physical systems.

\section{Related Work}
We first analyze the literature from the perspective of robot teleoperation before highlighting how we depart from the related work. Then, we look at the critical challenges associated with achieving a reliable mobile robot teleoperation system with the AT framework from a wireless network standpoint.

Teleoperation of a robot refers to controlling a robot remotely via a controller device (e.g., a haptic controller for a manipulator or a gamepad controller for a mobile robot). Bilateral teleoperation is a well-researched topic, which refers to the architecture of two-way (simultaneous or synchronous) control between a Master and a Client robot \cite{ferre2007advances}. Several bilateral teleoperation control laws and protocols have been proposed to address the challenges of control stability, time-delayed networks etc. \cite{wirz2009bidirectional}.
For instance, in \cite{Franken2011}, the authors resolved the time-varying destabilization issue of the bilateral system caused by the relaxed grasp of the user, hard contacts, stiff control settings, or communication delays by splitting the control architecture into two layers. The top layer is responsible for implementing a strategy to achieve desired transparency, and the bottom layer ensures that no virtual energy is generated. Separate communication channels connect the layers at the Client and the Master side. On the same lines, a recent work \cite{Zakerimanesh2019} introduced a control framework for bilateral teleoperation for a manipulator remote robot to ensure position tracking of the end-effector while carrying out sub-task control such as obstacle avoidance.  

While most of the work in bilateral teleoperation focuses on manipulators \cite{Davies2015}, researchers have also looked at applying this concept to mobile robots \cite{li2021dual,Shahzad2016,Vences-Jimenez2018,hou2015dynamic}. 
In \cite{Farkhatdinov2010}, a new feedback force rendering scheme for the teleoperation of mobile robots is proposed by adaptively tuning it based on range-based distance measured from the obstacles and time derivatives of those distances. The proposed methodology of variable force feedback gain performs significantly better than the conventional feedback schemes proposed in the literature. 
The authors in \cite{Kawai2010} proposed a control law for achieving bilateral teleoperation on a mobile robot, using a virtual Master robot that drives the real Client robot in a remote environment. 
The stability of the system is achieved by utilizing the scattering theory, and passivity-based control scheme \cite{li2021dual}. 
In another work \cite{Lee2011a}, researchers used a haptic teleoperation control framework for multiple Unmanned Aerial Vehicles (UAVs) to demonstrate how a single remote human user can stably teleoperate multiple distributed UAVs with some helpful haptic feedback over the Internet with varying delay, packet-loss, and multiple layered abstractions for control.
Authors of \cite{Bu2016} implemented a similar bilateral teleoperation technique by using the online generation of virtual fixtures for bilateral teleoperation based on intention recognition.
Similar to \cite{Franken2011} involving a single-Master-single-Client system, \cite{Minelli2019} puts forward a novel scheme for multi-Master-multi-Client teleoperation by decomposing the system into Master-Client pairs and implementing the standard duo; coordination strategy and transparency layer on them all the while using only two shared energy tasks between the Masters and the Clients. 
Similarly, \cite{Li2019b} proposed bilateral teleoperation of multi-robots by a single Master over a delayed network by scheduling communication. At each sampling instant, only one Client can transmit the information over the communication channel via scheduling protocols like Round Robin. 

As we can see, the great bulk of research is devoted to ensuring the stability of closed-loop systems with a human in the loop. These stability issues are tackled by time-domain passivity control or scattering theory. Sometimes the emphasis is placed on maintaining stability with model mismatches that comes into play. Contrary to these works, we differ by providing a new perspective to bilateral teleoperation \cite{tahir2020robot} and propose a novel priority-based switching (multiplexed) control for bilateral teleoperation achieving an open-loop synchronization between a Master and a Client robot embedded into our AT framework that can accommodate both human and an AI agent in the control loop.

We use an open-loop control method to reduce stability concerns and eliminate the requirement for model mismatch because the Master device is a mobile robot \cite{Kawai2010}. Secondly, the literature is keen primarily on devising alternate methods of force feedback calculation. The feedback imposed on a Master device has to be a haptic or tactile force that creates a sense of transparency for the human operator. 
In our work, we use predictive force (indicative of the obstacles ahead) to provide the Master robot with the essential cue to reroute its path for safer supervised navigation. 
Further, our framework is aided by the recent developments in remote computation technology. This enables the capabilities of a real-time algorithm adaptation and supervised control system using artificial intelligence (AI) or a human operator.

The key challenge here is to balance networking, computing, and control performances, especially since network connectivity is crucial and should be rapid and seamless for realizing a stable, high-performance teleoperation system. However, we cannot send all the sensor data to the remote device every time, and realistic wireless channels exhibit intermittent, low bandwidth, and unreliable transmission, mainly due to the mobility of the robot \cite{pandey2021empirical,parasuraman2014wireless}. 
While the computational performance such as CPU/Memory utilization is optimized through the use of remote computation \cite{7052418}, we must consider the network-related parameters such as the delay and throughput as they will have a significant impact on the control performance of the proposed system \cite{Li2020,Shahzad2016}.
Also, the need for predictive feedback that is not designed to be applied on a joystick is particularly challenging because much of the work in the literature focuses on remote teleoperation. Still, only a few have delved into the autonomous (with AI in the loop) bilateral teleoperation domain. 
Therefore, this paper aims to improve the supervised teleoperation task through remote offloading and achieve higher computation and network performances at the field robot compared to the conventional schemes of remote offloading and teleoperation.

\section{Background and Preliminaries}
Our approach uses remote bidirectional teleoperation to allow a Master robot to help a Client robot move to a destination in an unknown environment using map-based path planning. The Master robot receives the predictive feedback force from the Client to determine the field environment. 

Specifically, through the proposed AT framework, we envision a teleoperation system that relies on an AI agent (or a human operator), such as an autonomous robot (Master) located at a cloud/remote resource, to remotely control another computationally less intelligent robot (Client) through a network. The Client follows the Master's path from the remote location while delivering raw sensory data back to the AI at the Master. The AI uses this sensor data to control the Master, which is reflected at the Client, allowing it to execute complicated tasks like obstacle avoidance and navigation at the remote location while simultaneously offloading computing loads from the Client. 
Preliminaries to the main aspects of our proposed scheme are discussed below. 

\begin{figure}[t]
\centering
    \begin{center}
    \includegraphics[width=.98\columnwidth]{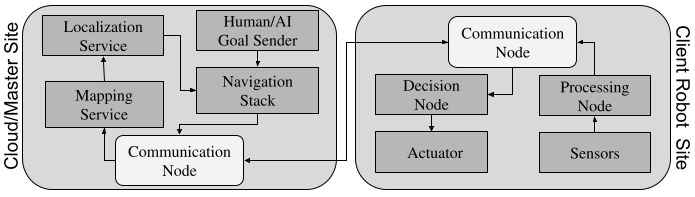}
    \caption{Architecture of a remotely-offloaded mobile robot navigation task example.}
    \label{fig:architecture}
    \vspace{-4mm}
\end{center}
\end{figure}

\subsection{Navigation and Path Planning}
To navigate across an unknown environment while avoiding collisions, a robot has to typically solve the Simultaneous Localization and Mapping (SLAM) problem  \cite{SMITH1988435}. SLAM provides an estimate of the map of the robot's environment while localizing the robot in relation to that map, resulting in a joint estimation problem. Different probabilistic formulations are used in approximating the pose of the robot and map, e.g. Extended Kalman Filters (EKF) \cite{jensfelt2001active}, Markov \cite {fox1999markov}, Monte Carlo Particle Filter (MCL) \cite{thrun2001robust}, and Rao-Blackwellization technique \cite{montemerlo2002fastslam}. 
Appendix \ref{sec:pathplanning-appendix} provides further details on path planning using the SLAM map, specifically the Dynamic window approach (DWA) \cite{fox1997dynamic} planner used in this study.

\subsection{Remote Computation}
\label{sec:remotecomputation}

The idea of remote computing and cloud robotics has sparked a lot of curiosity in the recent decade. Localization, navigation, perception, and mapping-related tasks can be offloaded to a remote workstation or a cloud because of the limited computing capability of the mobile robots \cite{galambos2020cloud}. 
The basic theme of a remote computation setup is shown in Fig. \ref{fig:architecture}.
For instance, in \cite{7052418}, the authors describe the viability of using the cloud for offloading low-level and intensive computing tasks such as the vision-based navigation assistance of a mobile service robot. In \cite{benavidez2015cloud}, the authors propose offloading the SLAM processing to clouds with cloud-based deployment. Another work in \cite{zhang2018cloud} suggests a cloud-based collaborative Visual SLAM system for transferring vast amounts of data. 

In \cite{arumugam2010davinci}, the authors use a cloud service running the FastSLAM algorithm, demonstrating a significant improvement in execution time compared to running the SLAM on the robot. Another paper \cite{miratabzadeh2016cloud} proposes a cloud-based (i.e., remote computing) architecture for large-scale autonomous robots that utilize three subsystems, including the middleware subsystem, background task subsystem, and control subsystem. These jobs may need immediate or deferred processing.

In our work, we use remote computation by outsourcing the tasks of mapping and localization to the remote mobile robot, Master, with better computational capability to perform these tasks based on the sensor data received from the less capable on-field robot (Client). 
The Client robot on the field navigates to the desired goal position through this process.

\subsection{Force Feedback}
\label{sec:reactive}

In classical bilateral mobile robot teleoperation, the human operator exerts a forward force on the Master device (manipulator or a controller) that causes displacement, which is transmitted to the Client robot to mimic that movement. If the Client robot has force or torque sensors, it sends a reflective force back to the Master. This reaction force comes from its interaction with its environment, which enters into the input torque of the Master, and the Client is said to be \textit{bilaterally} controlled.

With the use of force feedback, the operator's control over the robot's motion can be significantly improved \cite{hou2015dynamic}. For the feedback force, haptic or visual devices are typically introduced into the system at the Master to enhance the human operator's sense of telepresence by presenting the interaction force with the environment acting on the Client robot \cite{owen2013haptic}. These devices offer a new mode of perception and transparency of the remote environment by enhancing a user's depth perception as well as environmental awareness with obstacle detection. 

In a predictive force feedback system, a system of controllers is used based on the mobile robot's distance to the obstacles in the environment such that the force is perceived (predicted) before actual contact with the environment \cite{liu2017predictive}.
With the help of this information, the operator avoids navigating into the proximity of the obstacles and enhances the telepresence of the remote user. 
As a robot's distance from nearby obstacles decreases, a force reciprocal to the distance from the obstacle is computed and delivered through the haptic device, which the user perceives as a force and steers the robot toward a smooth and obstacle-free route. Such a system is highly effective in search and rescue mission scenarios \cite{linda2010fuzzy}. 
We introduce this predictive force into our teleoperation system to create the bilateral loop, where the feedback augmentation reflects the proximity of obstacles in the steering direction.

\begin{figure*}[t]
    \centering
    \includegraphics[width=.90\textwidth]{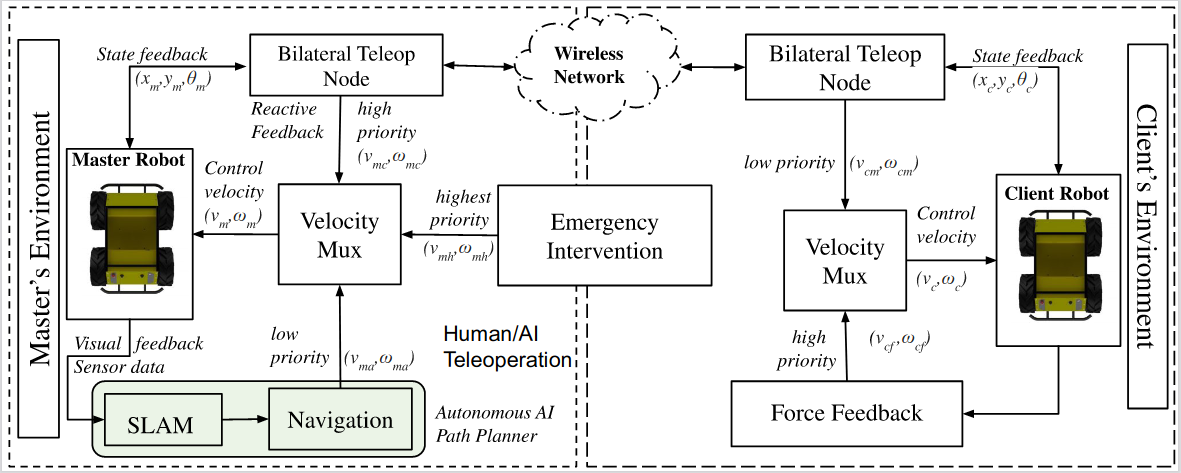}
    \caption{Proposed priority-based bilateral teleoperation mechanism to enable Master-Client coupling for AI-driven navigation.}
    \label{fig:mainfig}
    \vspace{-2mm}
\end{figure*} 

\section{Proposed Methodology} 
\label{sec:proposedmethod}

In our work, a teleoperation scheme for a single-Master-single-Client system is set up based on the proposed analog twin framework, where both the Master and Client robots are coupled through a bilateral control strategy. The method addresses the problem of situation transparency by implementing reflective commands on the Client and enhances better navigation by prioritizing path planning and teleoperation. We evaluate computational offloading on the remote agent that functions as an AI controller via a wireless network.

We introduce prioritized bilateral teleoperation through a Master robot situated on the remote server, which provides map-based path planning to the Client. The Client, in turn, provides reactive feedback for corrective navigation achieving overall goal point precision and obstacle avoidance over a wireless network. Unlike conventional approaches, this strictly predictive feedback force is provided to an autonomous Master robot, based on expected contact with the environment, rather than an impact force traditionally applied to a haptic device operated by a human. While executing autonomous map-based navigation directed by a path planner, the Master robot assures rerouting of the Client around obstructions by leveraging feedback.

Suppose there is a human in the loop (e.g., to provide emergency interventions), the Master robot functions as a physical surrogate of the remote agent on the control site. The predictive force applied to the Master comes from the Client and serves for improved awareness for the human operator. 

In a nutshell, the proposed methodology comprises a \textit{priority-based bilateral control} to teleoperate/navigate a Client robot towards a specified goal through an autonomous Master robot remotely over a wireless network. The Client robot subscribes to the Master's state to replicate the Master's motion at the field site. However, the Client robot comes equipped with an added layer of safe navigation for reactive obstacle avoidance provided through a fast, low-intensive \textit{predictive force feedback} computation conveyed to the Master robot through a high-priority channel. We employ a velocity multiplexer (MUX) on both sides for switching control between different control velocities based on priority. This MUX-based architecture generates open-loop control, which ensures system stability. An overview of our idea is shown in Fig.~\ref{fig:mainfig}. Below, we describe the key components of our method.

\subsection{Predictive Force Computation}
Consider the wheeled robot, with control inputs consisting of linear velocity $v_{m}$ and the angular velocity $\omega_{m}$ of the Master robot are derived from its position ($x_{m},y_{m}$) and rotation $\theta_{m}$. The kinematics model of a wheeled robot is given in Eq.~\eqref{eqn:equation3}.

\begin{equation}
\label{eqn:equation3}
 \begin{bmatrix}
  x_{m} \\
  y_{m} \\
  \theta_{m} \\
\end{bmatrix} = 
\begin{bmatrix}
    \cos(\theta)_{m} & 0 \\
    \sin(\theta)_{m} & 0 \\
      0            & 1  
\end{bmatrix}
\begin{bmatrix}
  V_{m} \\
  \omega_{m} \\
\end{bmatrix} 
\end{equation}
The control of the Client robot is based on the Master's linear and angular velocities, as shown in Eq.~\eqref{eqn:maincontrol}.
\begin{equation}
\label{eqn:maincontrol}
\begin{pmatrix} V_{c}\\
  \omega_{c} \\
\end{pmatrix}=
\begin{pmatrix}
  k_{V} & 0 \\
  0 & k_{\omega}
\end{pmatrix}
\begin{pmatrix}
  V_{m} \\
  \omega_{m}
\end{pmatrix}
\end{equation}
where $k_{V}$ and $k_{\omega}$ are scaling constants to account for dissimilar kinematics or scale (robot dimensions). 
Our proposed predictive force delivered via the Client robot is based on obstacle range information. This force is calculated based on the measured distances of the mobile robot to the obstacles coming in through the range values from the LIDAR data, demonstrated through Eq.~\eqref{eqn:equation5}.
\begin{equation}
\label{eqn:equation5}
{\Delta = R_{o} - R = (\delta_{1} \quad  \delta_{i} \quad   ...  \quad  \delta_{n} )} ,
\end{equation}
where, $i$ = 1....$n$ and $n$ is a number of ranges (or directions) in the sensor data. $R$ is a polar vector that represents the distance of the robot to the obstacles $r_{i}$. While $R_{o}$ is a vector of distance from the obstacle $r_{oi}$ which generates the predictive force. If the distance from the closest obstacle from the robot at any given instance of time becomes less than $r_{oi}$, it generates the force reflected back to the Master robot via feedback force. $\Delta$ defines the difference between $R_{o}$ and $R$. In short, the following basic law governs the principle of predictive force applied. 
\begin{eqnarray}
\label{eqn:equation6}
f_{i} = 
\begin{cases}
  \frac{k_{i}}{\delta_{i}},&\; r_{i} <  r_{oi} \\ 
  1, &\; r_i = r_{oi} \\
  0, &\; r_{i}\ge r_{oi}
\end{cases} \\
\label{eqn:equation7}
F = (f_{i} \quad f_{2}  \quad .... \quad f_{n}) \\
F \leftsquigarrow f_{i} \label{eqn:equation8} 
\end{eqnarray}
Here, $f_{i}$ is a predictive force calculated inversely proportional to the obstacle distance $\delta_{i}$ measured via $i^{th}$ sensor (or $i^{th}$ direction) with a multiplication gain (weight) $k_{i}$ for that direction such that $f_i \geq 1 \forall r_i \geq r_{oi}$. $k_{i}$ is usually kept constant, but can also be a variable depending on the scenario.

\subsection{Priority-based Control Strategy}
Similar robots, either real or simulated, are considered on the Master and Client sides. 
The Master robot is controlled by AI input in a low-priority channel, while the Client robot is controlled using Eq.~\eqref{eqn:maincontrol} on a low-priority channel.
Both are physically and logically separate entities connected via a communication channel with time-varying delay characteristics. The Master acts as an enabler for the resource-constrained Client by allowing it to "offload" computational and storage-intensive activities like mapping, localization, and navigation to the Master.

The proposed priority-based Master-Client coupling performs concurrent bidirectional teleoperation on a low-priority queue through a velocity multiplexer (MUX) on both sides. A MUX arbitrates incoming commands velocity messages from several topics, allowing only one channel (topic) at a time to publish, based on priorities. 
The bilateral coupling works by first enabling the Master robot to subscribe to the velocity commands from the Human/AI on a low-priority channel while, in parallel, the Client subscribes to the Master's velocity on a low-priority channel. Second, both agents subscribe to the reactive feedback force on a high-priority channel. This force coupling with the Master, based on the predictive force in Eq.~\eqref{eqn:equation8}, aids the Client in navigating its local environment while avoiding obstacles.

Specifically, we apply the below reactive force feedback $(V_r,\omega_r)$ on the high-priority MUX channels of both the Client and the Master robots. 
This feedback is designed in a way to turn the robot around the nearest obstacle direction. 
\begin{eqnarray}
\label{eqn:predictive-velocity}
  V_r & = & f_{safe}(d) \nonumber \\  
  \omega_{r} &  = & f(F) = 
\begin{cases}
    \omega_{turn},&\; max(f_i) \geq f_{th} \\   
    0, &\; otherwise
\end{cases} \\
\begin{pmatrix} V_{m}\\
  \omega_{m} \\
\end{pmatrix} & =
\begin{pmatrix} V_{c}\\
  \omega_{c} \\
\end{pmatrix} & =
\begin{pmatrix}
  k_{V} & 0 \\
  0 & k_{\omega}
\end{pmatrix}
\begin{pmatrix}
  V_{r} \\
  \omega_{r}
\end{pmatrix}
\label{eqn:ATcontrol}
\end{eqnarray}
Here, $\omega_{turn}$ is a fixed angular velocity, $f_{safe}(d)$ is a function to apply safe linear velocity, and $f_{th}$ is a force threshold.

The above force feedback mechanism (Eq.~\eqref{eqn:predictive-velocity}) is only taken as an example to show the effectiveness of the AT framework. However, it is possible to apply complex controllers to this framework to achieve AI verification-related objectives, e.g, we can design functions that can provide disturbances in the system to simulate uncertainty in the sensor data or the environment to verify the robustness of an AI controller (at the Master). Further, we can consider the Master robot as the AI teleoperator employing SLAM-based navigation to steer itself and the Client robot to a specified target in the Master's frame of reference. The readjustment of the route after applying force feedback on the Master robot is necessary for safer navigation around a highly complicated environment on the Client's side.

\begin{figure*}[t]
    \centering
    \begin{subfigure}{.29\textwidth}
        \includegraphics[width=\textwidth]{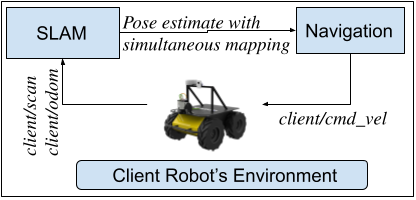}
        \caption{Case 0 (Onboard Computation).}
        \label{fig:case0}  
    \end{subfigure}
    \begin{subfigure}{.31\textwidth}
        \includegraphics[width=\textwidth]{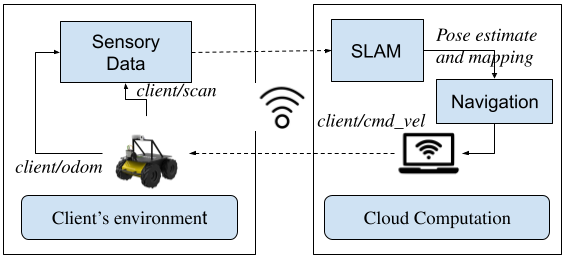}
        \caption{Case 1 (Remote Computation).}
        \label{fig:case1}  
    \end{subfigure} 
    \begin{subfigure}{.37\textwidth}
        \includegraphics[width=\textwidth]{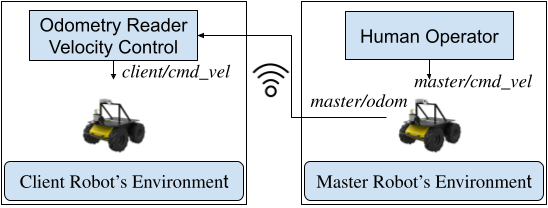}
        \caption{Case 2 (Manual Teleopration).}
        \label{fig:case2}  
    \end{subfigure}
    \caption{Thematic representation of the experiment cases analyzed in this experimental study. Here, 'odom', 'scan', and 'cmd\_vel' are the ROS topics for odometry, LIDAR scan, and command velocity (control) information.}
    \label{fig:thematicrep}
    \vspace{-2mm}
\end{figure*} 
\subsubsection{System Stability}
In a classical closed-loop bilateral teleoperation system, the Master maintains constant control over the Client by position and velocity control inputs, followed by feedback via corrective position. This may render the overall system to become unstable under time-delayed information flow, necessitating the use of additional control schemes to ensure system stability \cite{Li2020}. 

However, our proposed control in Eq.~\eqref{eqn:maincontrol} (low priority) and ~\eqref{eqn:ATcontrol} (high priority) creates a reliable and modular system. The velocity multiplexer and the channel priorities will ensure that the system is in an open loop at a given time ensuring the stability of the system. at both the robots, as shown in Fig. \ref{fig:mainfig}. At any instance of time, the Client only subscribes to the Master's velocity in an open-loop manner (Eq.~\eqref{eqn:maincontrol}). When an obstruction is nearby, the Client and Master robots switch to Eq.~\eqref{eqn:ATcontrol} to avert a potential collision. 
Once the possibility of a collision is averted, the regular navigation controller takes over, thereby exhibiting a switching-based control. 

This strategy also keeps the AI planner aware of the change in state and responds accordingly.
Specifically, because the AI is reactive in our navigation task, it does not require a high-precision control loop, which allows us to accommodate the AT framework ensuring stability and safety. For instance, research in \cite{shull2009open} found that open-loop architectures are more preferred as they avoid stability issues, transmitting motion commands while allowing any force feedback only via sensory substitution. Therefore, this finding is valid for both human and AI operators controlling a robot.

\subsubsection{Advantages of Analog Twin Framework}
The AT framework has the below advantages over a digital twin framework.
\begin{itemize}
    \item AT can be added to an already established digital twin through the proposed priority-based synchronization; 
    \item AT provides reliable and stable synchronization between the physical and remote environments; 
    \item AT provides additional storage and computational power through remote offloading for the physical system with reduced network-induced latency; 
    \item AT provides an avenue in closing the simulation-to-real gap in the design and verification of robot algorithms.
\end{itemize}

\begin{figure*}[ht]
    \centering
    \includegraphics[width=.98\textwidth]{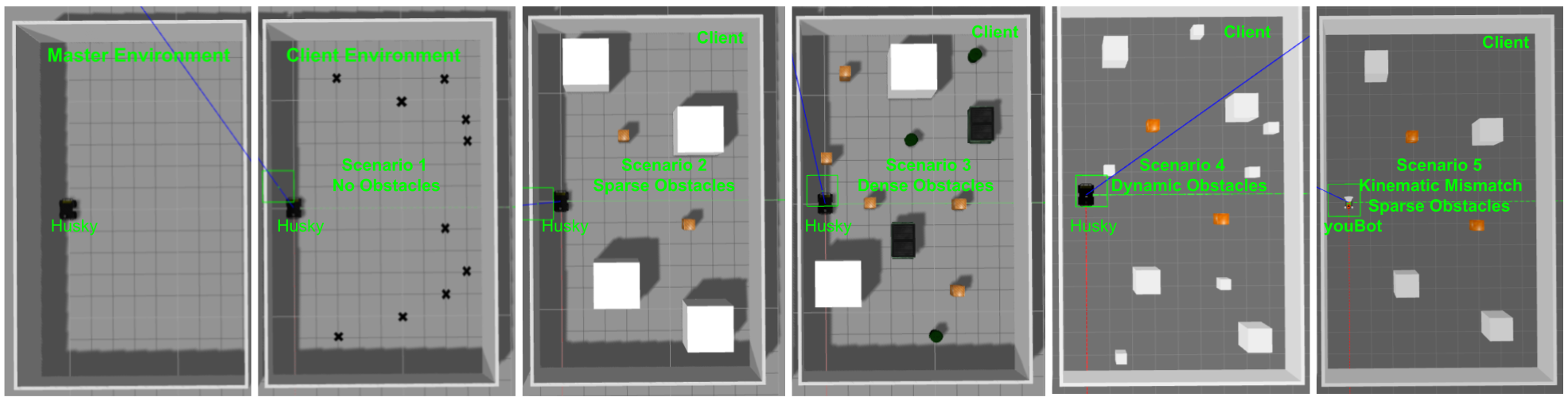}
    \caption{Experiment setup of the simulations in ROS Gazebo. The Master environment always has no obstacles, while the Client side has no obstacles (Scenario 1), sparse obstacles (Scenario 2, 4, and 5), dense obstacles (Scenario 3), and additional dynamic obstacles (Scenario 4). The goal locations are shown with a cross in Scenario 1, which are the same for all scenarios.}
    \label{fig:simulation-scenarios}  
\end{figure*} 
\section{Experimental Analysis}
\label{sec:expcases}

In our work, we implement the methods on the popular Robot Operating Systems (ROS) software framework \cite{quigley2009ros}, providing open-source software tools and libraries for both simulation and real-world hardware experiments. 
We utilize the \textit{move\_base} navigation package in ROS (library), which uses the Dynamic Window Approach (DWA) \cite{fox1997dynamic} for the local (base) planner and the $A^*$ \cite{le2018modified} for the global planner to navigate towards the goal. To learn grid maps from the robot's laser range scanner and wheel encoder sensors, we use the Gmapping SLAM package \cite{grisetti2007improved}, which is based on the Rao-Blackwellization technique with a particle filter in our simulation experiments. We use the Hector SLAM \cite{kohlbrecher2013hector}, which is based on an Extended Kalman Filter approach, in our hardware experiments. This is due to the compatibility with the Husky mobile robot platform used in ROS Gazebo simulations and the Turtlebot robot platform used in hardware experiment trials.
In both simulations and actual hardware experiments, we implement the following experimental cases (strategies) to compare the performances.

\paragraph{Case 0 [Baseline: onboard computing] - standalone Client robot with SLAM running onboard)} We test our system against the conventional scheme of an AI path planner driving a standalone mobile robot (see Fig.~\ref{fig:case0}). 
The robot creates the map of the environment using SLAM while using it to navigate toward the goal location.
The performance of this setup is used as a baseline for comparison with the proposed strategy.

\paragraph{Case 1 [Baseline: remote computing] - Client robot with conventional remote computation)} We implement a typical "remote-brained robot" scenario where physically and logically, the robot's body and its "brain" are separated, and remote intelligence is applied to the robot as shown in Fig.~\ref{fig:case1}. The goal is given to the Client robot through the output of the SLAM mapper and path planner over the network. 
No Master robot is employed, but a remote computer/laptop is used on the Master side. This case enables the Client robot to remotely offload its computationally extensive tasks like SLAM, path planning, and navigation to a cloud-based server or agent. 

\paragraph{Case 2 [No Feedback] Master-Client coupled with human teleoperation)} In this case, we experiment with a Master robot that is teleoperated in its environment and the Client robot subscribes to the odometry data from the Master in an open-loop manner as shown in Fig.~\ref{fig:case2}. The Master robot is steered by a human user through a joystick towards the desired target, with visibility to only the Master's environment whilst the Client side is obscured to the user. As the Master moves towards the goal point, the Client robot remotely follows the Master robot through the velocity commands read over the network and replicates them in its own environment. Force feedback (or bilateral coupling) is not applied in this case and no SLAM mapping is used on either side. This provides an ideal scenario for comprehending the location tracking inaccuracy between the Master and Client over a network. 

\paragraph{Case 3 [Proposed AT] Master-Client coupling with prioritized force feedback)} We implement our proposed robot-robot synchronization methodology (Fig.~\ref{fig:mainfig}) to test the accuracy, delay, and time elapsed of the navigation task. According to our proposed approach, the Client robot offloads the computational overhead of mapping and navigation to the Master side and accepts the velocity commands from the Master robot on a low-priority multiplexer channel, and navigates its environment through the motion commands received from the Master. The mapping and navigation are performed by the Master of its environment, while the Client implements the velocity computed by the Master robot. The Master tries to arrive at the goal position based on the information from the map, while the Client simply follows the Master's odometry commands remotely and gets to the goal position. The Client reacts to the obstacles in the environment and sends predictive force feedback reacting to the contacts from the environment. Based on the feedback, the Master performs a corrective action on its trajectory to arrive at the goal position.

\section{Simulation Experiments}
\label{sec:experiments}

For testing the effectiveness of the proposed method, we perform experiments using ROS-based Gazebo 3D physics-engine robot simulations with a Clearpath Husky robot (differential drive) model as both a Master and a Client in two identically similar experiment rooms of size 17m x 8m. The robots are simulated in two laptops connected via a real Wi-Fi network, which is shown to have an average of 10ms delay in the network. We achieve overriding of velocity commands through a velocity multiplexer in the ROS package, $twist\_mux$, both on the Master and Client sides. The multiplexer on Husky is configured via a $.yaml$ file by prioritizing the velocity topics that reload at run time. The Client can detect nearby obstacles within the $r_{oi}$=0.5 m range. The parameters $f_{th}$=1 and $\omega_{turn}=0.5 \frac{rad}{s}$ set in the simulations.

\subsection{Experiment Scenarios}
\label{sec:scenarios}
We test these cases in three different scenarios, as shown in Fig.~\ref{fig:simulation-scenarios}. In all these scenarios, the Master robot's environment is kept constant (empty room), meanwhile varying the Client's environment to be an empty room, a room with sparse obstacles, and a room with dense obstacles for scenarios 1, 2, and 3, respectively. For each scenario and every case, we conduct at least five trials, and the results show the variations across the trials for every performance metric measured. For each trial, a different goal location was set, ranging from 6 m to 10 m in the distance from the origin (robot's start) (see Fig.~\ref{fig:simulation-scenarios}). These goal locations are the same across the cases and scenarios.

Furthermore, we add two more scenarios based on the second scenario (sparse obstacles) to verify the robustness of the approach in the presence of dynamic obstacles (Scenario 4), and the flexibility in the case that the Master and Client robots have dissimilar kinematics (Scenario 5), where the Master is a Husky robot with differential drive kinematics and the Client is a youBot robot with omnidirectional kinematics. More details on these two scenarios are given in Appendix \ref{sec:flexibilityandrobustness}.

\subsection{Performance Metrics}
\label{sec:metrics}
This study covers the assessment of the below measures to evaluate the system's performance.
\begin{itemize}
\item Navigation Task Performance 
    \begin{itemize}
        \item Goal accuracy (m): Absolute difference between the final robot position and the desired goal position
        \item Tracking error (m): Average of the absolute difference between the instantaneous positions of the Master and the Client robots
        \item Efficiency (s): The time elapsed to reach a goal point by the Client robot
    \end{itemize}
\item Network Performance
    \begin{itemize}
        \item Network throughput (Mbps): The average data rate at which the Master-Client machines communicate
        \item Network latency (s): The average delay with which the Master and Client machines communicate
    \end{itemize}
\item Computing Performance
    \begin{itemize}
        \item CPU utilization (\%): The average CPU load (measured every 5 minutes) at the Client robot/machine during the experiments
        \item Memory utilization (Mb): The average RAM memory used by the Client robot during the experiments
    \end{itemize}
\end{itemize}

\begin{figure*}[ht]
\centering
\begin{subfigure}{.32\textwidth}
\centering
\includegraphics[width=\linewidth]{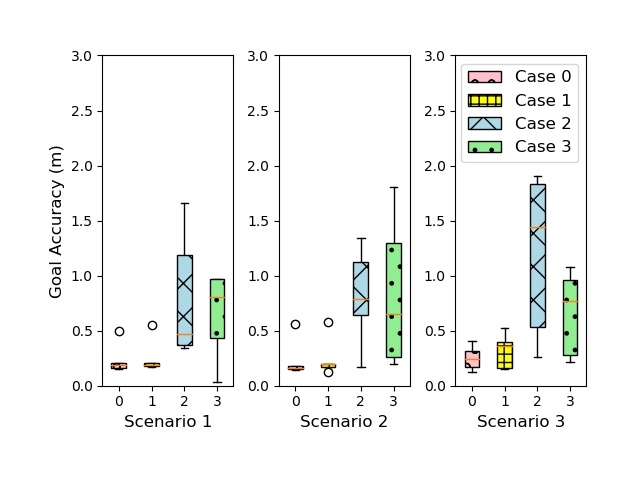}
\vspace{-8mm}
\caption{End Goal Accuracy}
\label{fig:goal_error_sim}
\end{subfigure}
\begin{subfigure}{.32\textwidth}
\centering
\includegraphics[width=\linewidth]{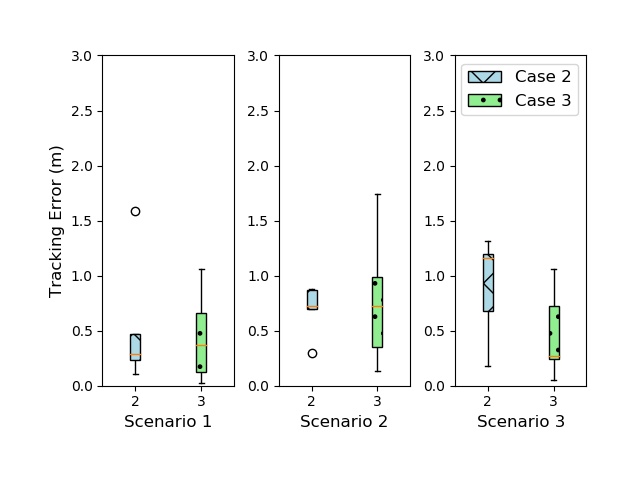}
\vspace{-8mm}
\caption{Master-Client Tracking error}
\label{fig:tracking_error_sim}
\end{subfigure}
\begin{subfigure}{.32\textwidth}
\centering
\includegraphics[width=\linewidth]{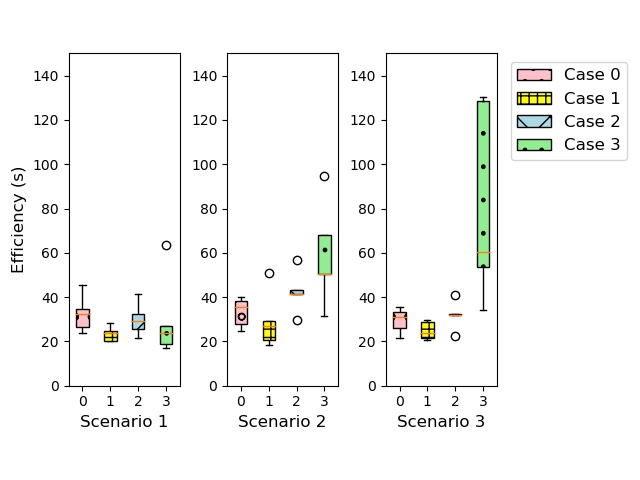}
\vspace{-8mm}
\caption{Task Efficiency}
\label{fig:efficiency_sim}
\end{subfigure}
\caption{Navigation task performance - end goal accuracy, tracking error, and efficiency in simulation experiments.}
\vspace{-4mm}
\label{fig:simulation-performance-plots}
\end{figure*}

\begin{figure*}[ht]
\centering
\begin{subfigure}{.25\textwidth}
\centering
\includegraphics[width=\linewidth]{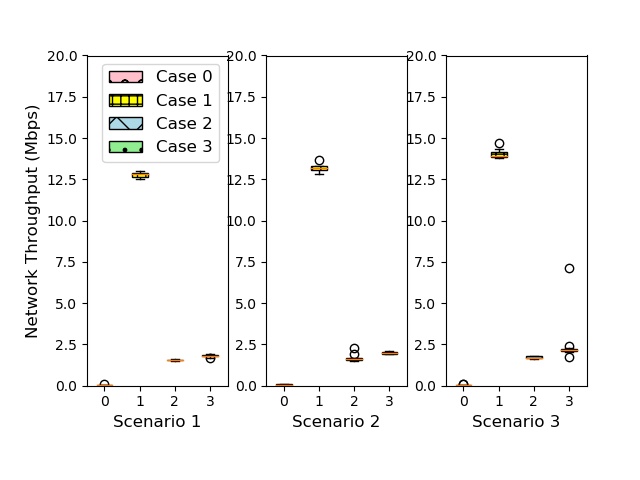}
\vspace{-8mm}
\caption{Network throughput}
\label{fig:network_throughput_sim}
\end{subfigure}
\begin{subfigure}{.25\textwidth}
\centering
\includegraphics[width=\linewidth]{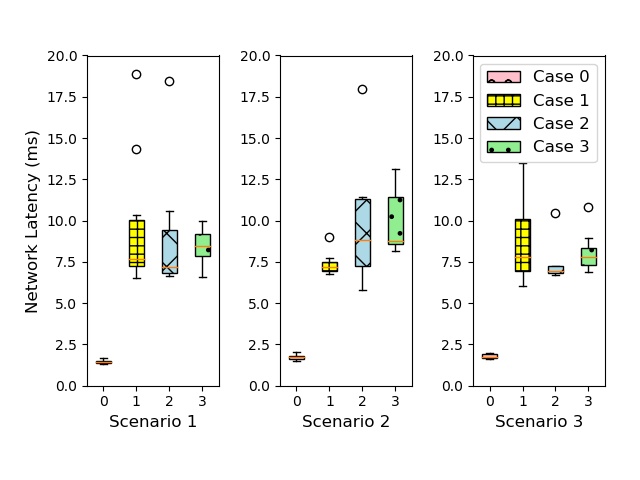}
\vspace{-8mm}
\caption{Network latency}
\label{fig:network_latency_sim}
\end{subfigure}
\begin{subfigure}{.23\textwidth}
\centering
\includegraphics[width=\linewidth]{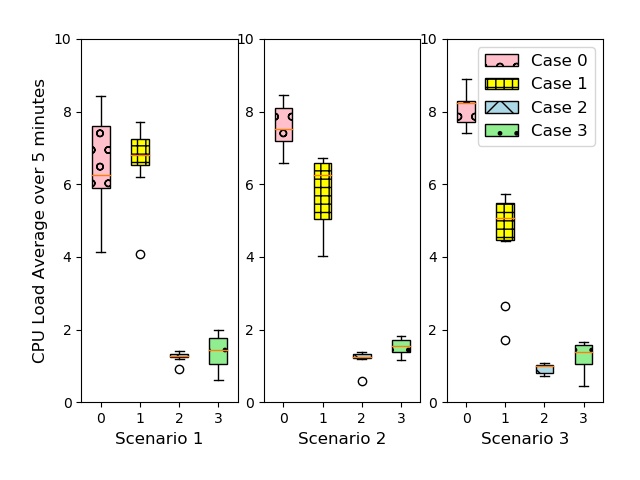}
\vspace{-8mm}
\caption{CPU utilization}
\label{fig:cpu_utilization_sim}
\end{subfigure}
\begin{subfigure}{.25\textwidth}
\centering
\includegraphics[width=\linewidth]{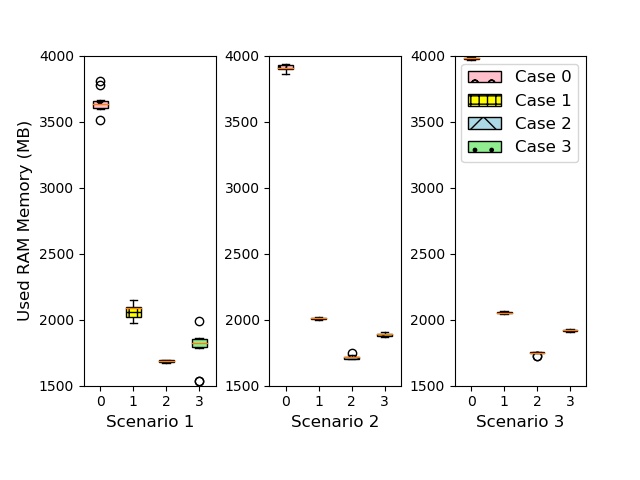}
\vspace{-8mm}
\caption{Memory Usage}
\label{fig:memory-usage_sim}
\end{subfigure}
\caption{Network and computing performance results in simulation experiments.}
\label{fig:simulation-network-computing-plots}
\end{figure*}

As the study focuses on remote computation, we are more interested in the wireless network quality and performance for remote teleoperation \cite{parasuraman2017new}. The majority of data from the Client site is delivered to the Master site, which is in charge of navigation. For instance, Case 1 transmits full raw sensor data to the Master, while Case 2 and Case 3 exchange only partial information.
Since sustaining communication is critical to our proposed approach, network latency must be continuously monitored. The Master robot sends the data, and the Client receives and re-sends new information obtained from the change in its environment. In addition, we analyze how network performance impacts computational offloading and if there is a significant lag or performance degradation while the data is being offloaded or had been offloaded.

We use our \textit{ROS Network Analysis} package, hosted in a public GitHub repository\footnote{\url{https://github.com/herolab-uga/ros-network-analysis}} for measuring network performance metrics through ROS nodes. For instance, the ping action implemented through ROS actionlib is used to calculate the latency or network delay. At the same time, we extract the network throughput reported by the Wi-Fi device driver.

\subsection{Results and discussions}
We analyze the results from the perspective of each metric and then discuss the flexibility and robustness.

\paragraph{End Goal Accuracy}
The results of the experiments performed across different cases are summarized in Fig.~\ref{fig:goal_error_sim}.
We test our strategy against the baseline scheme, Case 0, in which the robot is a standalone robot that maps and navigates its own environment without the assistance of any Master or autonomous agent.
The findings obtained in an open-world environment (scenario 1) reveal that our proposed strategy (Case 3) has a significantly controlled goal error, amounting to no more than 1m error, better than the teleoperation tracking (Case 2).
Because of autonomous navigation, Cases 0 and 1 show the best results in task performance metrics.

For scenario 2, the mean value of end goal error for Case 3 is slightly lower than that of Case 2. But, the accuracy of Case 3 is significantly better than Case 2 in scenario 3, where the obstacles get denser. 
Navigation without feedback becomes difficult due to obstacles blinded in the Master environment, as clearly seen by the performance of Case 2 getting worse over successive scenarios. However, our proposed strategy manages to keep the end goal accuracy under control because of the force feedback to the AI planner.

\paragraph{Tracking Error}
The tracking performance between the Master and the Client robot is shown in Fig.~\ref{fig:tracking_error_sim} for Cases 2, and 3 since only these two cases involve having a Master and a Client robot. 
The performance of Case 2 in scenario 1 is the baseline for this analysis as that provides us the best case tracking possible with a velocity subscriber at the Client. 
Tracking is challenged by the number of obstacles, as seen by the deviation between trials in Case 2 results. 
However, the feedback mechanism applied in Case 3 helps maintain consistent performance across the scenarios and significantly improves the tracking performance compared to Case 2.

\paragraph{Task Efficiency}
According to the data obtained shown in Fig.~\ref{fig:efficiency_sim}, the baseline cases show consistent efficiency across scenarios as the AI planner uses the SLAM map. 
It is interesting to note that Case 1 (remote computation) efficiency is better than Case 0 (onboard computation), demonstrating CPU load's impact on task performance.
In scenario 1, the proposed method performs at par with the baseline cases of autonomous navigation and remote offloading because of no impact on the obstacles. 
However, as expected, the efficiency of the proposed method is significantly challenged by the presence of obstacles. This is because dense obstacles necessitate a stronger and more persistent reaction to the force feedback loop in Case 3. An improved feedback function in Eq.~\ref{eqn:predictive-velocity} could be able to help overcome this issue. 

\paragraph{Network Throughput}
Performance comparison with respect to the throughput experienced by all the cases except the baseline (Case 0, which does not use a wireless network) is summarized in Fig.~\ref{fig:network_throughput_sim}. It determines the number of network packets exchanged between the Master and the Client robots at any given time. Throughput is averaged through the length of the trial. As expected, the throughput calculated for Case 1 would always be greater than the other cases, as it involves full data offloading to a remote workstation. 
As we can see, the proposed method provides at least a 5x reduction in network throughput than Case 1 while still performing core SLAM computation remotely, thanks to the feedback and synchronization strategy. Case 3's throughput is only marginally greater than Case 2, where the major data being shared is the Master robot states. This explains that the addition of reactive force in an open-loop manner does not increase data requirement considerably, significantly helping the system with better task performance while ensuring system stability.

\paragraph{Network Latency}
The network latency experienced by the system in all scenarios is shown in Fig. \ref{fig:network_latency_sim}. 
As the baseline case doesn't involve using a network, being a standalone robot, it experienced no network latency. 
Latency is relatively consistent and comparable in most cases, as the wireless network connecting the Master and Client machine is within a lab environment, ensuring the stability of the network and the control loop.
The variations in network latency can be suppressed by sending fewer data via the network, as demonstrated by the results of scenario 3 comparing Case 1 (sharing all data) and Case 3 (sharing partial data), indicating the usefulness of the suggested technique.

\paragraph{CPU and Memory Utilization}
The Client robot's computing performance (CPU load and memory use) across all cases in different scenarios is compared in Figs. \ref{fig:cpu_utilization_sim} and ~\ref{fig:memory-usage_sim}, respectively. 
As expected, Case 0 has the highest CPU load, and memory use since all computations are done locally on the Client robot. 
Again, in Case 1, because the CPU is tasked with continuous network transmission to offload the data to the Master side, while the memory is unused for computations, the Client robot shows a high CPU load but low memory use.
Case 2 has the lowest CPU load and memory usage because the Client robot is only tasked with replicating the Master robot, requiring the least computing resources.
The proposed Case 3 provides comparable performance to Case 2 since it optimally balances both the computing load and remote computation. We can observe up to 3x better CPU load and more than 2x better RAM use in Case 3 compared to the baseline cases.

\begin{figure}[t]
\centering
\vspace{-5mm}
\includegraphics[width=\linewidth, height=6cm]{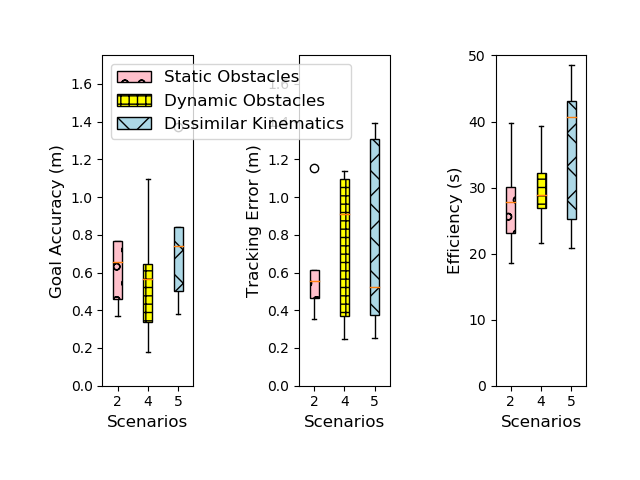}
\vspace{-10mm}
\caption{Comparison of Case 3's task performances in the presence of static obstacles (Scenario 2), dynamic obstacles (Scenario 4), and dissimilar kinematics (Scenario 5).}
\label{fig:flexibility}
\end{figure}

\subsubsection{Flexibility and Robustness}
To analyze the robustness of the approach (Case 3) with dynamic obstacles in Scenario 4, we look at Fig.~\ref{fig:flexibility}. Here, the goal accuracy was not affected by the introduction of moving obstacles, and a reasonable synchronization performance between Master-Client was maintained throughout all goal points tested. Also, the task efficiency did not suffer in Scenario 4 compared to Scenario 2. Overall, similar task efficiency was observed regardless of the spontaneity of the obstacles encountered. As expected, through the indicated results, we can derive that satisfied task performance can be obtained under the proposed framework in a highly dynamic setting. 

To analyze the flexibility of the approach with dissimilar kinematics at the Master and Client side, we compare Scenario 5 in Fig.~\ref{fig:simulation-scenarios} to the baseline Scenario 2, where the kinematics are the same under the same obstacle environment. Our results shown in Fig.~\ref{fig:flexibility} indicate that our proposed AT-aided feedback produced comparable task performance outcomes in terms of end-goal accuracy, Master-Client tracking, and efficiency in kinematically mismatched Master-Client pairs. While the end goal and tracking error remained well within the average of $\pm{0.75}$m for all goal points tested, the task efficiency of the dissimilar Husky-youBot pair experienced a slight decline from the similar robots Husky-Husky setup. We believe this is mostly due to imperfect synchronization amongst robot drivers with varied reaction times. 
The results indicate that the proposed scheme of remote collaboration between dissimilar Master-Client mobile robots is viable.

\begin{figure*}[t]
\centering
\begin{subfigure}{.24\textwidth}
\centering
\includegraphics[width=\linewidth]{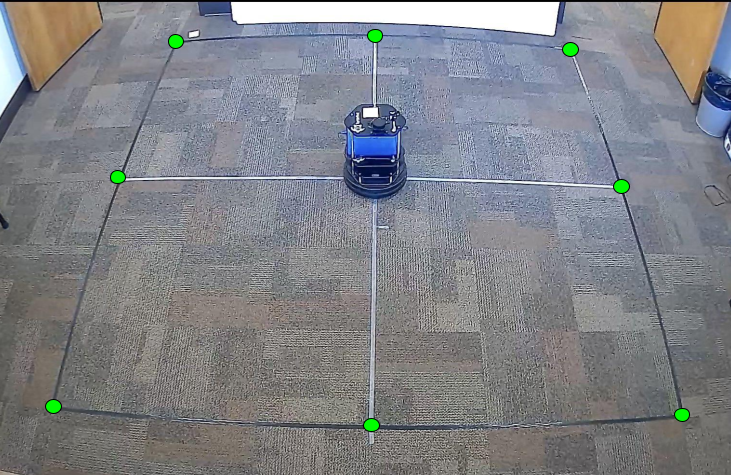}
\caption{Master Room showing the goal locations in green dots}
\label{fig:Masterroom}
\end{subfigure}
\begin{subfigure}{.29\textwidth}
\centering
\includegraphics[width=\linewidth]{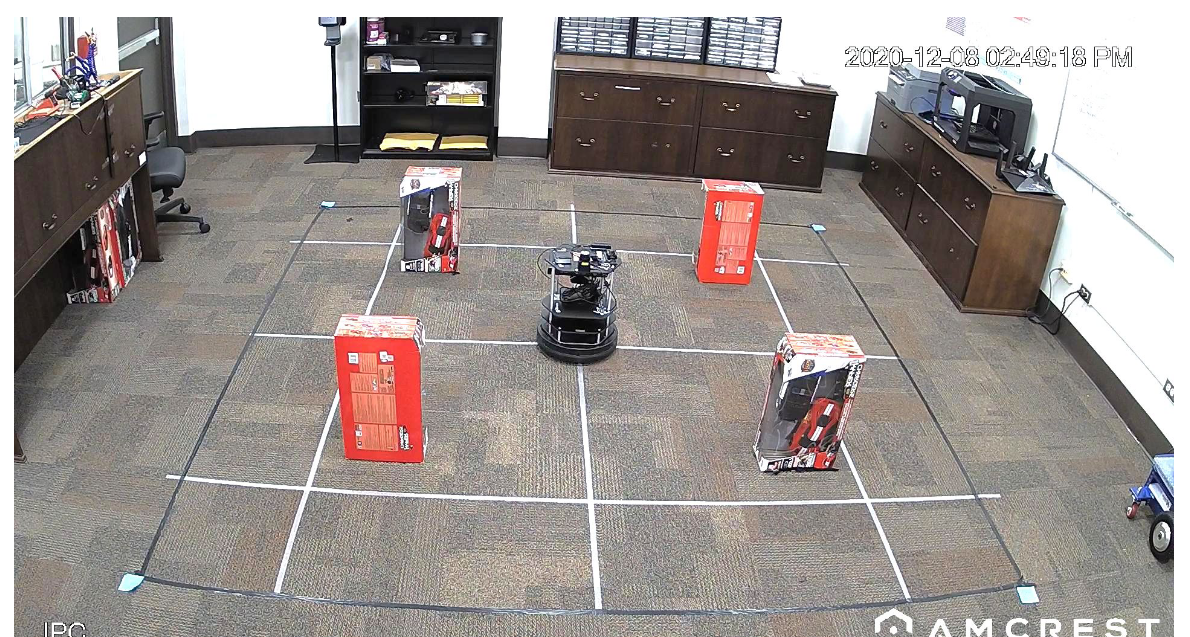}
\caption{Client Room showing the obstacle (red boxes) placements}
\label{fig:Clientroom}
\end{subfigure}
\begin{subfigure}{.21\textwidth}
\centering
\includegraphics[width=\linewidth]{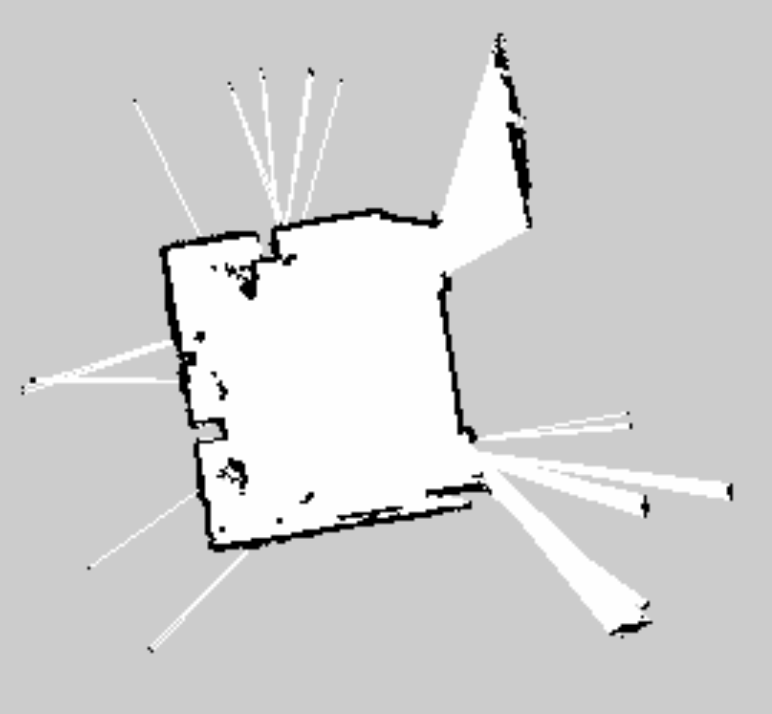}
\caption{Master room SLAM map}
\label{fig:MasterSLAM}
\end{subfigure}
\begin{subfigure}{.22\textwidth}
\centering
\includegraphics[width=\linewidth]{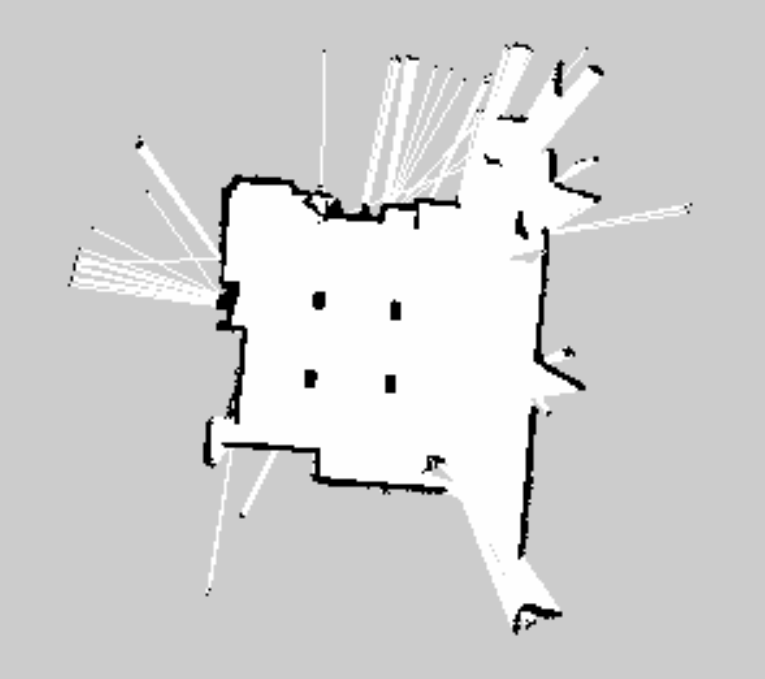}
\caption{Client room SLAM map}
\label{fig:ClientSLAM}
\end{subfigure}
\caption{Experiment setup for the hardware experiments with the SLAM maps generated by the Client and Master robots.}
\label{fig:hardwaremaps}
\end{figure*}

\subsubsection{Summary and Limitations}

The proposed solution has achieved a balance between task performance, networking, and computing requirements, but at a cost of reducing task efficiency, especially when the environment is complex or unpredictable. Therefore, this aspect needs to be further investigated.
Our system has some flexibility and resilience due to the use of variable feedback open-loop control, which proved useful even in a complicated setting with no human intervention. 
Although there is a possibility of congestion and an increased load on the Master robot, this mechanism is designed to keep the Master's load at the minimum by running the odometry reader node on the Client robot and by the use of a multiplexer that prevents deadlock and prioritizes the incoming velocity commands.
It is worth noting that the above analyses are exclusively from the Client robot's perspective, as it is the target of interest. We assume the Master robot is in a secure location without restriction to computing resources, enabling integration with Digital Twin.

\section{Real-world Robot Hardware Experiments}

To evaluate the real-time expediency of the proposed scheme, we test the same four experimental cases as in simulations (see Sec.~\ref{sec:expcases}). We use two identical mobile robots, Turtlebot 2e's, in our laboratory environment. Both the robots are connected via a Wi-Fi network (same as in simulations) and are located approx 20 meters apart in two separate experiment rooms, mimicking the Master and Client sides. Each room is marked with a 3x3m sub-square with rectangular obstacles. Eight trials are conducted for each case, with goals ranging within a ($\pm{1.5},\pm{1.5})$ offset from the origin (center). 
The hardware trials consist of one scenario (environment with four obstacles) as shown in Fig.~\ref{fig:hardwaremaps}. Here, the robot always starts from the center, and the eight goal locations are located at the corners of the 3x3m area, as shown in Fig.~\ref{fig:Masterroom}.

The Master robot has a 2D-LIDAR LDS-01 laser scanner, and the Client robot is equipped with a Hokuyo URG-04LX-UG01 2D-LIDAR laser range scanner as sensors used to create a SLAM map of the environment.  
Both robots are connected to computers running Ubuntu 18.04 with ROS melodic.

\emph{Performance Metrics}: In addition to the metrics outlined in Sec.~\ref{sec:metrics}, we include another metric called \textit{throughput loss}, which is the difference between the throughput obtained between the Master and the Client side. It is indicative of the packet loss in the network to  estimate how much of the offloaded data was lost in the network between the one-way (Cases 1 and 2) and two-way exchange (Case 3) of the data between the machines.

\emph{Obstacle placement}: 
The Client environment has obstacles laid as in Fig.~\ref{fig:Clientroom}. The SLAM map outputs are shown for the Master and Client rooms in Figs.~\ref{fig:MasterSLAM} and \ref{fig:ClientSLAM}, respectively.
The Master room has no obstacles for all cases, except Case 2, where it is intentionally kept this way to mimic the obstacles set up in the Master room for guiding human teleoperation. This way a reasonable route to goal points can be manually realized around the obstacles (without an AI planner).

\begin{figure*}[ht]
\centering
\begin{subfigure}{.24\textwidth}
\centering
\includegraphics[width=\linewidth]{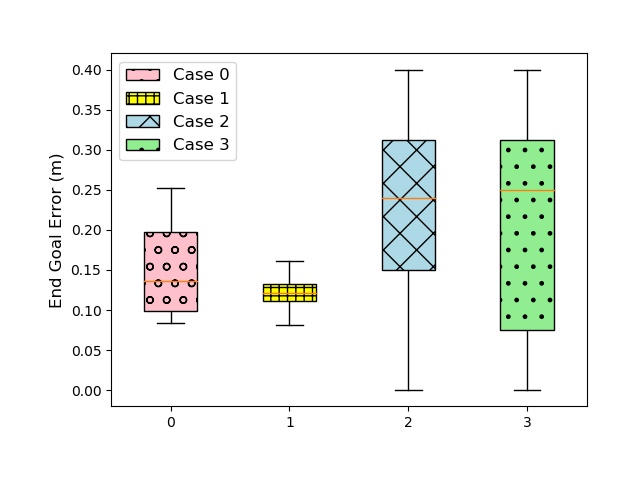}
\caption{End Goal Accuracy}
\label{fig:hardware-goalaccuracy-plot}
\end{subfigure}
\begin{subfigure}{.24\textwidth}
\centering
\includegraphics[width=\linewidth]{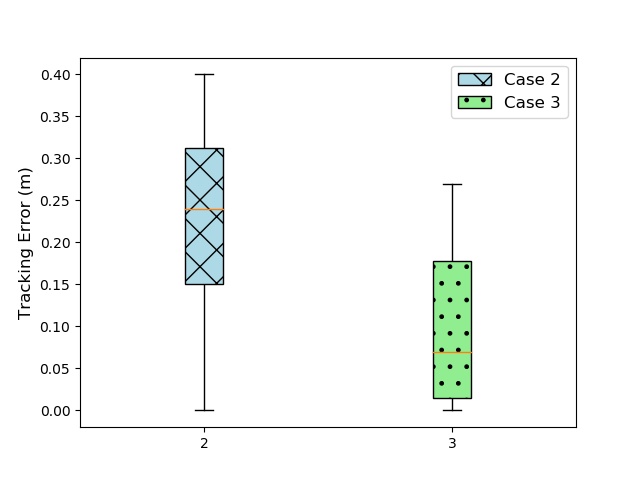}
\caption{Master-Client Tracking Error}
\label{fig:hardware-trackingaccuracy-plot}
\end{subfigure}
\begin{subfigure}{.24\textwidth}
\centering
\includegraphics[width=\linewidth]{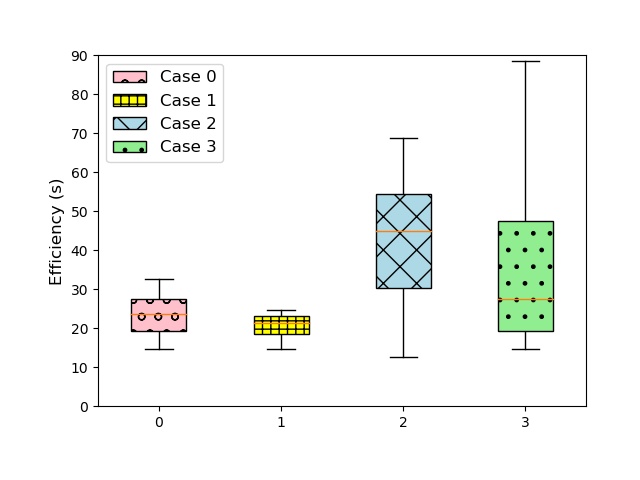}
\caption{Task Efficiency}
\label{fig:hardware-efficiency-plot}
\end{subfigure}
\begin{subfigure}{.24\textwidth}
\centering
\includegraphics[width=\linewidth]{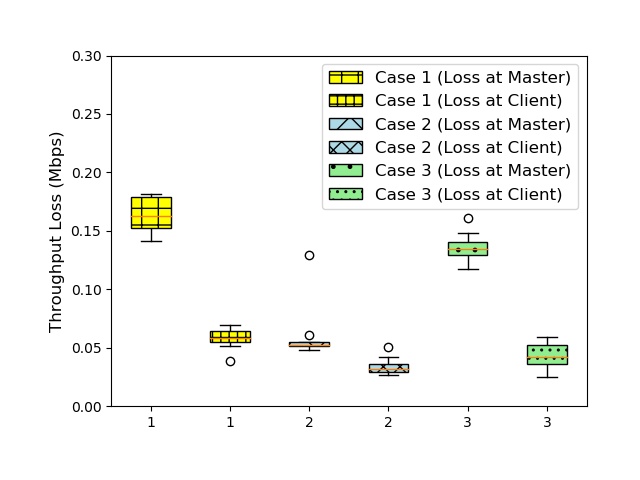}
\caption{Throughput loss}
\label{fig:hardware-throughput-loss}
\end{subfigure}
\caption{Task performance (End goal error, tracking error and efficiency) and throughput loss in hardware experiments.}
\vspace{-4mm}
\label{fig:hardware-goal-plots}
\end{figure*}

\begin{figure*}[ht]
\centering
\begin{subfigure}{.24\textwidth}
\centering
\includegraphics[width=\linewidth]{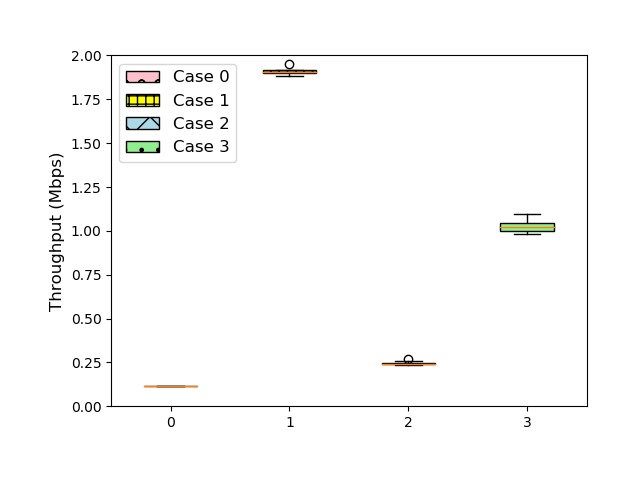}
\caption{Network Throughput}
\label{fig:hardware-throughput-plot}
\end{subfigure}
\begin{subfigure}{.24\textwidth}
\centering
\includegraphics[width=\linewidth]{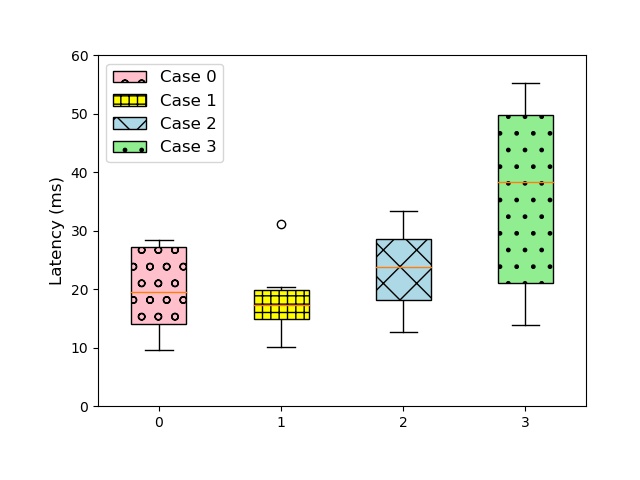}
\caption{Network Latency}
\label{fig:hardware-delay-plot}
\end{subfigure}
\begin{subfigure}{.24\textwidth}
\centering
\includegraphics[width=\linewidth]{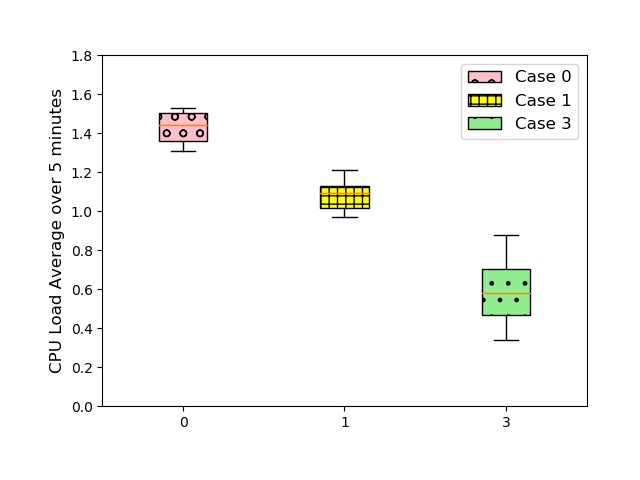}
\caption{CPU Utilization}
\label{fig:hardware-cpu}
\end{subfigure}
\begin{subfigure}{.24\textwidth}
\centering
\includegraphics[width=\linewidth]{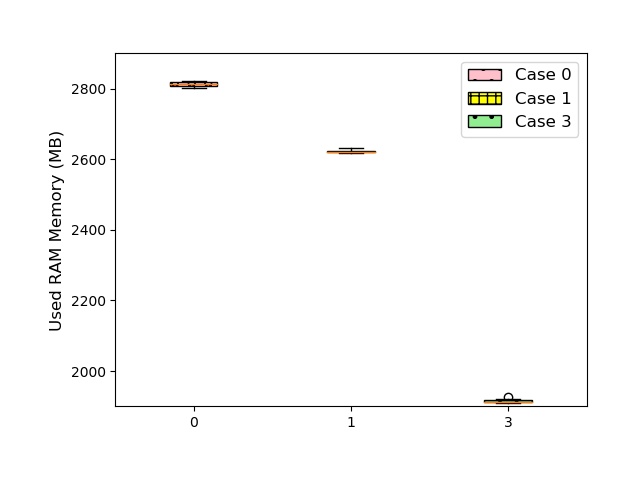}
\caption{Memory Usage}
\label{fig:hardware-memory-plot}
\end{subfigure}
\caption{Network and computing performance parameters in the real-world robot (hardware) experiments.}
\label{fig:hardware-computing-plots}
\end{figure*}

\subsection{Results and Discussions}
We discuss the results obtained in the hardware experiments below and compare them to the simulation results.

\paragraph{End Goal Accuracy}
The end goal accuracy result is shown in Fig.~\ref{fig:hardware-goalaccuracy-plot}. The proposed approach performed at par with Case 2, showing a minimal influence in reaching the end goal location due to the force feedback. It is worth noting that the performance of remote computation (Case 1) is better than that of the onboard computation (Case 0), possibly due to the higher computational resources spent on the robot.

\paragraph{Tracking Error}
Fig.~\ref{fig:hardware-trackingaccuracy-plot} shows the comparison of tracking error between Master and Client robots with and without force feedback. 
Through bilateral teleoperation of the mobile robots, we aim for the Client robot to follow a similar trajectory as planned by the Master robot for its environment, with feedback cues indicative of the Client environment to move the robot to the goal. 
Case 3 shows significantly less error in mimicking the Master side, demonstrating the advantages posed by the proposed priority-based feedback strategy.

\paragraph{Task Efficiency} 
We observe through the data collected in Fig.~\ref{fig:hardware-efficiency-plot} that it takes more time on average to arrive at the goal position when using a bilateral teleoperation system with predictive force as expected. Nevertheless, the proposed system efficiency is comparable to the remote offloading scenario without transmitting all sensor data. 

\paragraph{Throughput loss}
Throughput loss data presented in Fig \ref{fig:hardware-throughput-loss} shows the results for Cases 1 to 3 where there is a data exchange between the Master and the Client. Although this metric is not a direct measure of packet loss, it indicates how packets are retransmitted and handled at the two machines for the same data communication.
Since Case 1 involves heavy data offloading from the Client, the throughput loss of this case is the highest among all, as anticipated. 
Also, we expect that the throughput loss at the Master is generally higher than the loss at the Client since the Master sends its state (position) to the Client continuously at a constant rate and receives sensor or feedback data from the Client. This is seen in the outcome of all the cases.
Case 2 has the least amount of data to transmit, showing the lowest loss. The proposed strategy (Case 3) shows a balanced loss compared to Cases 1 and 2. This demonstrates the optimum and effective use of the network.

\paragraph{Network Throughput}  
The throughput utilized by all the cases in hardware experiments is summarized in Fig.~\ref{fig:hardware-throughput-plot}. 
As expected, the throughput of Case 1 is the highest due to offloading computationally expensive tasks like mapping and navigation and getting data back to the Client. Between Cases 2 and 3, data exchange is lowest for Case 2 as teleoperating the Master robot drives the Client robot without exchanging any information from the Client, while Case 3 does entail two-way communication owing to feedback requiring larger yet balanced throughput. 
Throughput in Case 3 is between 1-1.15 Mbps due to two-way data transferred and received, unlike Case 1. 
Case 3 shows around a 2x reduction in network throughput requirement, allowing higher scalability of operations using the proposed strategy.

\paragraph{Network Latency} 
We observe more network latency for Case 3, as seen in Fig.~\ref{fig:hardware-delay-plot}. Network latency can lead to a broader tracking error, although with the exception of a few spikes for our testing trials, this latency is under control. Note that for all cases, we run the \textit{roscore} ROS node (which is responsible for establishing communication between the machines) at the Master machine, regardless of whether sensor/state data is exchanged (including the baseline Case 0). This is to obtain a realistic baseline for the minimum network latency for handshake communication with a remote machine in the same network. 
We observe an average latency of around 40ms for Case 3, about 2x higher than the baseline (Case 0).
Nevertheless, no significant performance degradation or tracking error is observed due to this delay.

\paragraph{CPU and Memory utilization} 
Compared to the baseline case, The CPU utilization drops by around 57\% for Case 3 compared to Case 0, as depicted in Fig.~\ref{fig:hardware-cpu}. 
The proposed strategy effectively balances between performing all the computation itself (Case 0) and offloading all the computational tasks to the remote machine (Case 1).
Memory utilization data is shown in Fig.~\ref{fig:hardware-memory-plot} follows a similar trend as in the CPU load metric. We observed a significant drop in Case 3's RAM usage compared to the baseline case. This is due to the offloading of the computationally hungry tasks which becomes the liability of the Master robot in a bilateral and collaborative teleoperation system. 
Note that we omitted presenting Case 2's data in these figures, as there was another process running on the Client robot during Case 2 trials that compromised our computing metrics measurement. 

The computing metrics prove that collaborative offloading leads to low computation requirements at the Client robot by leveraging on the communications network and the computational resources offered by the remote robot. 
Moreover, computation offloading prevents processing deadlocks at the Client-side and improves performance and efficiency.
However, the reliance on onboard computation decreases but leads to more trust in the network conditions, requiring a robust, reliable, and efficient communication network.

\section{Conclusion} 
\label{sec:conclusion}
We proposed an analog twin framework using a Master-Client collaboration architecture for synchronizing robots. The proposed approach can help simulate new algorithms and innovations for collaborative AI and real-time verification of algorithms. To illustrate the proposed framework, a novel priority-based bilateral teleoperation strategy is presented and tested for a human-teleoperated and AI-supervised navigation task. This strategy uses computational offloading for resource-constrained mobile robots that lack computational capacity. 

Our strategy is verified for effectiveness and flexibility through extensive simulations and real-world robot experiments. We analyzed the performance metrics from the perspective of task (tracking and goal accuracy and efficiency), networking (throughput and delay), and computation (memory and CPU usage). Our strategy outperformed conventional strategies like onboard computation, remote computation, and manual teleoperation schemes. The results demonstrate reductions in network and computing requirements without compromising task performance, signifying reliable AT collaboration. 

In general, similar trends are observed in both hardware and simulated cases for efficiency, throughput, and network latency.
However, a significant difference in the size of the robot's workspace created certain dissimilarities between the results from simulation and hardware studies. Moreover, in simulations, computing resources also consider the processing of the graphical visualization of the simulated robots.
In our future work, we will conduct experiments on continuous navigation tasks that could shed more light on the benefits and limitations of the proposed AT-aided collaboration framework.

We believe that the results of the proposed idea of bilateral teleoperation through a computationally efficient analog twin will aid in the research and development of novel human or AI in the loop robotics and Cyber-Physical Systems.


\clearpage
\appendices
\section{Map-based Path Planning}
\label{sec:pathplanning-appendix}
To navigate across an unknown environment while avoiding collisions, a robot has to typically solve the Simultaneous Localization and Mapping (SLAM) problem  \cite{SMITH1988435}. SLAM requires the robot to estimate a map of its environment while localizing itself in relation to that map, resulting in a joint estimation problem. Different probabilistic formulations are used in approximating the pose of the robot and map, e.g. Extended Kalman Filters (EKF) \cite{jensfelt2001active}, Markov \cite {fox1999markov}, Monte Carlo Particle Filter (MCL) \cite{thrun2001robust}, and Rao-Blackwellization technique \cite{montemerlo2002fastslam}. For instance, the widely-used Gmapping ROS package employs a Particle Filter based approach, while the Hector SLAM ROS package employs an Extended Kalman Filter based approach.

The most common algorithm employed to create static and dynamic maps is based on occupancy grid techniques. In such methods, occupancy probability is defined for each cell in both maps. The occupancy probability is based on the confidence of whether a respective cell is occupied or not. An inverse observation model is applied to determine occupancy probability on both static and dynamic maps to be either high or low based on the occupancy states in the previous maps and the current sensor information. The accuracy of this map improves over time in a structured environment as the robot explores its surroundings and continuously localizes itself with respect to landmarks or obstacles in the environment \cite{burgard2007mobile}. 

One of the most essential technologies in mobile robots is path planning. Path planning's major objective is to determine the smoothest path from the starting point to the destination point in a constrained environment, with or without obstacles, based on the map information generated by the SLAM algorithm. Path planning is divided into global path planning and local path planning. Local path planning requires relative location and obstacle avoidance, whereas global path planning requires information from global localization data. 

The global planner, such as the $A^*$ planner \cite{le2018modified} requires that the robot operates with prior knowledge of the surroundings through the SLAM map, which is put into the robot path planning before it begins moving. It specifies the route to take from the start to the endpoint. Obstacle avoidance is the task of local path planning. It allows the robot to stop before hitting an obstacle and performs an optimization that calculates the shortest, unobstructed path to the goal. It also stops the robot at the goal position and performs a recovery rotation. 

As we focus on the obstacle avoidance strategy in supervised teleoperation, we detail the local planner used in our work.
Dynamic window approach (DWA) \cite{fox1997dynamic} is a well-known local planner taking care of obstacle avoidance. DWA takes into account the dynamic and kinematic constraints of a mobile robot. It uses a cost map and a controller to issue velocity commands to the robot's mobile base. This method employs a grid map surrounding the robot and utilizes a value function to compute the velocities of traversing to the various points in the grid map. A velocity is supposed to be admissible if the robot is able to stop before it reaches the obstacle. 

\begin{equation}
\label{eqn:equation1}
\mathit{V_{a}} = \{ (\textit{v}, \omega) \; | \; \sqrt{\: 2 \: . \: dist(\textit{v},\omega) \; \cdot \; \dot{\textit{v}} \: } \; \wedge \; \omega \leq \sqrt{\: 2 \: . \: dist(\textit{v},\omega)  \: \cdot \: \dot{\omega} \:} 
\end{equation}
The Eq.~\eqref{eqn:equation1} represents ${V_{a}}$ as the set of velocities $(\textit{v}, \omega)$ that allows the robot to stop without colliding with an obstacle. The $\dot{\textit{v}}$ and $\dot{\omega}$ are the accelerations for breakage. Dynamic Window reduces the overall search space to the dynamic window, which has only those velocities that can be reached within the next time interval. A dynamic window is created to account for the limited accelerations exerted by the motors. Thus, the dynamic window $\textit{V}_{d}$ is defined as
\begin{equation}
\label{eqn:equation2}
\mathit{V_{d}} = \{ (\textit{v}, \omega) \; | \;  \textit{v}\varepsilon \:  [ \textit{V}_{a} - \; \dot{\textit{v}} \: \cdot \: \textit{t}] \; \wedge \; \omega_{\varepsilon} [ \omega_{a} \; - \; \dot{\omega} \; \cdot \; \textit{t}, \omega_{a} + \dot{\omega} \; \cdot \; t \}
\end{equation}
In Eq.~\eqref{eqn:equation2}, \textit{t} is the time interval during which velocities $\textit{v}$ and $\omega$ will be applied and (\textit{v}, $\omega$) is the actual velocity of the robot. 

For more robust obstacle avoidance, a fusion of sensor data from sonar, stereo vision, laser range finder, LIDAR and 3D depth sensors is generally used for obstacle detection. More advanced obstacle avoidance algorithms like artificial potential field method (PFM) \cite{Khatib85}, vector field histogram method (VFH) \cite{borenstein1991vector}, Bug algorithm \cite{lumelsky1987path}, Follow The Gap (FGM) method \cite{sezer2012novel} can be applied but to avoid computational overload on the Client side and to reduce network latency, in this study, only the LIDAR and inertial odometry sensors are utilized to perform SLAM and reactive obstacle avoidance.

In robot navigation, path planning is typically well integrated with perception algorithms in a closed loop. This invokes several computing tasks that the robot needs to handle such as the object recognition, perception, mapping, navigation, etc. SLAM \cite{gouveia2014computation}, SIFT-based object identification \cite{bistry2010cloud}, and object recognition and tracking \cite{nimmagadda2010real}, computer vision and perception \cite{zhang2013real} are all examples of potential remote offloading tasks because of their intensive computation requirement.


{\color{black}
\section{Flexibility and Robustness}
\label{sec:flexibilityandrobustness}
Some of the key challenges in mobile robot teleoperation in real environments include the master-client kinematic dissimilarity and environmental dynamics.
Therefore, to further verify the flexibility and robustness of the proposed method, we simulated a few scenarios and validate the proposed method against a baseline scenario of similar kinematics in a static obstacles environment (Scenario 2 in Fig. 5). 

\paragraph*{Robustness} First, we simulate a highly dynamic environment where moving obstacles in addition to the static obstacles were introduced in the experimental workspace so that the Client robot is exposed to a rapidly changing setup where the predictability of the static, structured environment is reduced and the robot is expected to provide timely feedback in response to an incoming moving obstacle. We introduced animated boxes moving in a wayward fashion on the Client's side of the simulated setup (Scenario 4 in Fig. 5) but this scenario can easily be extended to other moving robots sharing the workspace with the Client robot. For instance, we envision multiple mobile robots collaborating on shared tasks that require complementary computing and intelligence. 

We observed that the end goal accuracy was not affected by the introduction of moving obstacles as indicated in the results shown in Fig. 8 and a reasonable synchronization performance between Master-Client was maintained throughout the different goal points tested. The task efficiency did not suffer in the new scenario as well and overall same efficiency is observed regardless of the spontaneity of the obstacles encountered. As expected, through the indicated results, we can conclude that satisfied task performance can be obtained under the proposed framework in a highly dynamic setting. 

\paragraph*{Flexibility} Next, we look at the flexibility of the proposed approach with respect to using Master and Client robots with dissimilar kinematics. In a traditional bilateral teleoperation setting, the master device usually a joystick has a bounded workspace while the client mobile robot can/should cover an unbounded workspace. In our proposed framework, since the master and the client robots are both differential drive robots, a feasible tracking synchronization between the two robots was facilitated through velocity-velocity coupling. One may argue the applicability of the proposed analog twin framework on robots with kinematics mismatch. 

In order to test the feasibility of adopting alternative kinematics and flexibility of our approach, we propose the simulation results validated by using KUKA youBot as a Client robot; an omnidirectional vehicle platform based on Mecanum wheels. The Master robot is the Husky robot as used in all other scenarios. This setup is referred to as Scenario 5 in Fig. 5.

The omnidirectional aspect gives the KUKA youBot high maneuverability by allowing the vehicle to move in every direction of the plane, also termed as "holonomic" robots - robots capable of driving in any direction \cite{braunl2008omni}. The movement of the KUKA youBot base is caused by the four Mecanum wheels. A Jacobian matrix is presented to describe the motion of the Mecanum wheel. The Jacobian matrices can be combined using the Eq.~\eqref{eqn:youbot}. 
\begin{equation}
\label{eqn:youbot}
J_{youBot} = 
\begin{pmatrix}
    J_{C_{1}} & 0 & 0 & 0 \\
    0 & J_{C_{2}} & 0 & 0 \\
    0 & 0 & J_{C_{3}} & 0 \\
    0 & 0 & 0 & J_{C_{4}}
  \end{pmatrix}
\end{equation}
where $J_{C_{1}}$, $J_{C_{2}}$, $J_{C_{3}}$ and $J_{C_{4}}$ are Jacobian matrices of the four contact frames, at the axis of each wheel. 
We can define the Jacobian matrix for contact frame $i$ considering the orientation of the robot $\theta$ as: 
\begin{equation}
\label{eqn:equation9}
J_{C_{i}} = 
\begin{pmatrix}
    -R_{i}\sin(\theta)^R_{C_{i}} & r_{i}\sin(\theta^R_{C_{i}} + \eta_{i}) & -d^R_{C_{i}{y}} \\
    R_{i}\cos(\theta)^R_{C_{i}}  & -r_{i}\cos(\theta^R_{C_{i}} + \eta_{i}) & -d^R_{C_{i}{x}} \\
    0 & 0  & 1\\
  \end{pmatrix}
\end{equation}
Here, $R_i$ denotes the circumference of the primary wheel $i$ while $r_i$ specifies the circumference of the roller of wheel $i$. 

To make this equation simpler since the youBot have symmetric structure, the wheels and rollers all have the same circumference, therefore $R_{1}$=$R_{2}$=$R_{3}$=$R_{4}$=$R$ and $r_{1}$=$r_{2}$=$r_{3}$=$r_{4}$=$r$. The magnitude of the roller angles is also identical for wheels 1 and 3, while this angle is inverted for wheels 2 and 4. The KUKA youBot's Mecanum wheels are typical Swedish wheels. 

Therefore, $\eta_{1}$ = $\eta_{3}$ = $-45\degree$ and $\eta_{2}$ = $\eta_{4}$ = $45\degree$. 

The four wheels' Jacobian matrices are defined in the following equations. 
\begin{equation}
\label{eqn:equation10}
J_{C_{1}} = 
\begin{pmatrix}
    -R\sin(\theta) & r\sin(\theta - 45) &  T{y} \\
    R\cos(\theta)  & -r\cos(\theta - 45) & T_{x} \\
    0 & 0  & 1\\
  \end{pmatrix}
\end{equation}

\begin{equation}
\label{eqn:equation11}
J_{C_{2}} = 
\begin{pmatrix}
    -R\sin(\theta) & r\sin(\theta + 45) &  T{y} \\
    R\cos(\theta)  & -r\cos(\theta + 45) & -T_{x} \\
    0 & 0  & 1\\
  \end{pmatrix}
\end{equation}

\begin{equation}
\label{eqn:equation12}
J_{C_{3}} = 
\begin{pmatrix}
    -R\sin(\theta) & r\sin(\theta - 45) &  -T{y} \\
    R\cos(\theta)  & -r\cos(\theta - 45) & -T_{x} \\
    0 & 0  & 1\\
  \end{pmatrix}
\end{equation}

\begin{equation}
\label{eqn:equation13}
J_{C_{4}} = 
\begin{pmatrix}
    -R\sin(\theta) & r\sin(\theta + 45) &  -T{y} \\
    R\cos(\theta)  & -r\cos(\theta + 45) & T_{x} \\
    0 & 0  & 1\\
  \end{pmatrix}
\end{equation}
where, $T_x$ and $T_y$ represent the distance between the robot frame and the contact frame in x and y directions, respectively. 

Finally, the inverse kinematics of the youBot to drive the wheel angular velocities $\dot{\psi}_i$ based on the input (command) linear and angular velocities $(\dot{x},\dot{y},\omega)$ are give below \cite{li2019kinematic}.

\begin{equation}
\label{eqn:youbot_invkin}
\psi_{i} = J^T \begin{pmatrix}
    \dot{x} \\
    \dot{y} \\
    \dot{\theta} \\
  \end{pmatrix}
\end{equation}

Here, the main difference between the inverse kinematics of the youBot platform (omnidirectional kinematics) and that of the Husky platform (differential drive kinematics) is the linear command velocities. For differential drive, the $\dot{y} = 0$, while the youBot allows velocity in both x and y directions. We study the impact of this dissimilarity on task performance with our Analog Twin synchronization approach outlined in Case 3 of the proposed work. 

We tested the validity of our suggested scheme by coupling this holonomic Client (youbot robot) with a non-holonomic differential drive robot as a master robot (Husky robot) as illustrated with Scenario 5 in Fig. 8. Our results indicated that our proposed analog-twin bilateral teleoperation scheme with feedback produced comparable task performance outcomes in terms of end-goal accuracy, master-client tracking, and efficiency in kinematically mismatched Master-Client pairs. The end goal and tracking error remained well within the average of $\pm{0.75}$m for all goal points tested. The task efficiency of the dissimilar Husky (Master) - KUKA youBot (Client) however, experienced a slight decline from the similar robots Husky (Master) - Husky (Client) analog twin's case but that can be accounted for not having near perfect synchronization between the two dissimilar robots due to their device drivers implementing the low-level wheel control. The experimentation results indicated that the reliability, efficiency, and feasibility of our proposed scheme between dissimilar Master-Client mobile robots are viable and the remote collaboration between such robots is highly achievable with reasonable precision. 
}


\bibliography{Bilateral_teleoperation}
\bibliographystyle{IEEEtran}

\end{document}